\documentclass{article} 
\usepackage{iclr2018_workshop,times}
\usepackage{hyperref}
\usepackage{url}
\usepackage{pgfplots,capt-of}
\usepackage{accents}
\newcommand{\vectr}[1]{\accentset{\rightharpoonup}{#1}}
\newcommand{\vectl}[1]{\accentset{\leftharpoonup}{#1}}
  \usepackage{graphicx}

\title{SufiSent - Universal Sentence Representations Using Suffix Encodings}

\author{Siddhartha Brahma\\
IBM Research, Almaden, USA\\
\texttt{brahma@us.ibm.com} 
}

\begin{document}

\maketitle

\begin{abstract}

Computing universal  distributed representations of sentences is a fundamental task in natural language processing.
We propose a method to learn such representations by encoding the suffixes of word sequences in a sentence and training 
on the Stanford Natural Language Inference (SNLI) dataset. We demonstrate the effectiveness of our approach by evaluating it on
the SentEval benchmark, improving on existing approaches on several transfer tasks.
\end{abstract}

\section{Introduction}
In natural language processing, the use of distributed representations has become standard through the effective use of word embeddings. In a wide range of NLP tasks, it is beneficial to initialize the word embeddings with ones learnt from large text corpora like word2vec \cite{word2vec} or GLoVe \cite{glove} and tune them as a part of a target task e.g. text classification. It is therefore a natural question to ask whether such standardized representations of whole sentences that can be widely used in downstream tasks, is possible. 

There are two classes of approaches to this problem. Taking cue from word2vec, an unsupervised learning approach is taken by SkipThought \cite{skipthought} and FastSent \cite{hill2016}. More recently, the work of \cite{infersent} takes a supervised learning approach. They train a sentence encoding model on the  Stanford Natural Language Inference (SNLI) dataset  \cite{snli} and show that the learnt encoding transfers well to to a set of transfer tasks encapsulated in the SentEval benchmark. This is reminiscent of the approach taken by ImageNet \cite{imagenet} in the computer vision community.

One of the most effective ways of encoding a sentence $\mathbf{s}$ is to pass it through a recurrent neural network like a LSTM \cite{lstm} and use the last hidden state or a combination of the intermediate hidden states. Each intermediate state of a LSTM represents an encoding of a prefix of $\mathbf{s}$. In a bidirectional LSTM, an additional network is used to encode the prefixes of the reversed sequence of words. Although this is equivalent to encoding the suffixes of $\mathbf{s}$, the suffixes are encoded in  a direction reverse of the prefixes.

In this paper, we argue that encoding the suffixes of $\mathbf{s}$ in the forward direction can lead to better universal sentence representations. By Max-pooling the encodings of the prefixes and suffixes, we define a new sentence encoding that is trained on the SNLI dataset. We show through numerical experiments that the learned encodings improve upon  existing supervised approaches 
on the SentEval benchmark. We call our suffix based sentence encoding model \textsc{SufiSent}.

\section{Suffix Based Models}

Let $\mathbf{s}$ be a sentence with $n$ words. We will use $\mathbf{s}[i\colon j]$ to denote the sequence of words from $\mathbf{s}[i]$ to $\mathbf{s}[j]$, where $i$ maybe less than $j$.
Let $\vectr{L}_p$ represent a LSTM (or any other RNN) that encodes \textit{prefixes} of $\mathbf{s}$ in the forward direction. 
For the $i$-th word, we have
\begin{equation}
\vectr{h}_{p,i} = \vectr{L}_p(\mathbf{s}[1\colon i])
\end{equation}
Let $\vectr{L}_s$ represent a LSTM that encodes \emph{suffixes} of $\mathbf{s}$ in the forward direction. 
\begin{equation}
\vectr{h}_{s,i} = \vectr{L}_s(\mathbf{s}[i\colon n])
\end{equation}
Note that the $\vectr{h}_{p,i} $ can be computed in a single pass over $\mathbf{s}$, while computing $\vectr{h}_{s,i}$ needs a total of $n$ passes over progressively smaller suffixes of $\mathbf{s}$. 
As in bidirectional LSTMs, we also consider $\vectl{L}_p$ and $\vectl{L}_p$ that encodes the prefixes and suffixes of $\mathbf{s}$  in the backward  direction. 
\begin{equation}
\vectl{h}_{s,i} = \vectl{L}_s(\mathbf{s}[i\colon 1]), \hspace{.1in} \vectl{h}_{p,i} = \vectl{L}_p(\mathbf{s}[n\colon i]), 
\end{equation}
Note that $\vectr{h}_{p,i} $ encodes the same subsequence as $\vectl{h}_{s,i}$, but in different directions. See Fig.1(a) for a schematic illustration. Let $d$ be the dimension of the hidden state of each of  $\vectr{L}_p,  \vectr{L}_s,  \vectl{L}_p,  \vectl{L}_s$. We consider the following sentence encodings.
\begin{itemize}
\item \textsc{\textbf{SufiSent}} - We max-pool the $\vectr{h}_{p,i} $  over all $i\in [1:n]$ to obtain   $\vectr{h}_p$ and max-pool the $\vectr{h}_{s,i}$ to obtain $\vectr{h}_s$. Similarly, we obtain $\vectl{h}_p$  and  $\vectl{h}_s$ by max-pooling over $\vectl{h}_{p,i} $ and $\vectl{h}_{s,i}$ respectively. The final encoding is a concatenation of $\max(\vectr{h}_p,  \vectr{h}_s)$ and $\max(\vectl{h}_p,  \vectl{h}_s)$. The sentence encoding is of size $2d$. In contrast, a BiLSTM-Max model is a concatenation of $\vectr{h}_p$ and $\vectl{h}_p$.
\item \textsc{\textbf{SufiSent-Tied}} - This is same as above, but the weights of  $\vectr{L}_p$ and $\vectr{L}_s$ are shared or tied. Similarly, the weights of  $\vectl{L}_p$ and $\vectl{L}_s$ are tied. The sentence encoding is of size $2d$.
\item \textsc{\textbf{SufiSent-Cat}} - Similar to \textsc{SufiSent}, we compute $\vectr{h}_p,  \vectr{h}_s, \vectl{h}_p,  \vectl{h}_s$. The sentence encoding is the concatenation of these four vectors and is of size $4d$.
\item \textsc{\textbf{SufiSent-Cat-Tied}} - This is same as  \textsc{SufiSent-Cat}, except the weights of  $(\vectr{L}_p, \vectr{L}_s)$  and $(\vectl{L}_p, \vectl{L}_s)$ are tied. The size of the encoding is $4d$.
\end{itemize}
\begin{figure}[t]
\includegraphics[width=0.98\linewidth]{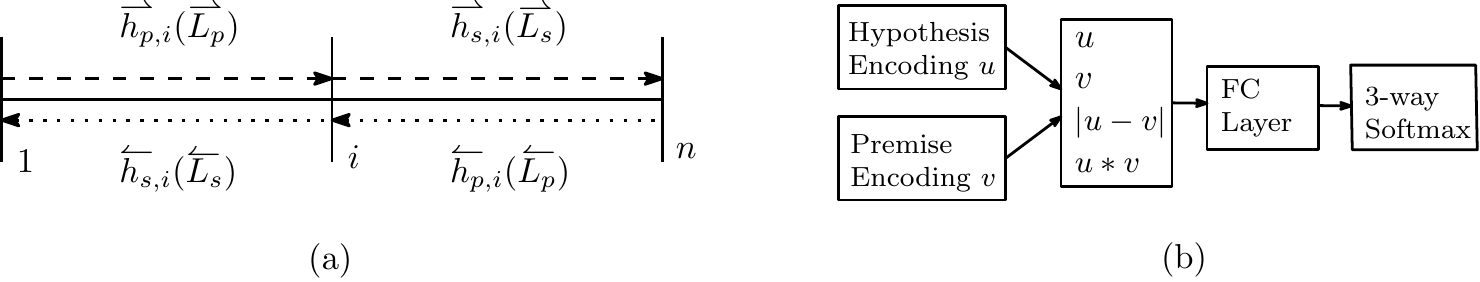}
\caption{(a) Schematics of \textsc{SufiSent} terms (b) Training architecture on SNLI dataset}
\end{figure}
We use the SNLI dataset as the supervised dataset to train the encodings. SNLI is a large scale labelled dataset consisting of pairs of sentences (premise and hypothesis) and each pair is labeled by one of three labels - entailment, contradiction and neutral. As shown in Fig. 1(b), for each of the \textsc{SufiSent} models, the encodings of the premise and hypothesis sentences are computed as $u$ and $v$. Following \cite{mou2016}, a feature vector consisting of $u$, $v$, $|u-v|$  and $u*v$  is fed into a fully connected layer(s), before computing the 3-way softmax in the classification layer. 

\section{Training and Results}
The encodings defined by \textsc{SufiSent} and \textsc{SufiSent-Tied} are trained on the SNLI dataset for the LSTM hidden dimensions $d \in \{256, 512, 1024, 2048\}$. The \textsc{SufiSent-Cat} and \textsc{SufiSent-Cat-Tied} encodings are trained for $d \in \{128, 256, 512, 1024\}$. This corresponds to sentence encoding dimensions of $512, 1024, 2048, 4096$ respectively. The FC layer has two layers of 512 dimensions each. For optimization, we use SGD with an initial learning rate of 0.1 which is decayed by 0.99 after every epoch or by 0.2 if there is a drop in the validation accuracy. Gradients are clipped to a maximum norm of 5.0.

We evaluate the sentence encodings using the SentEval benchmark \cite{infersent}. This benchmark consists of 6 text classification tasks (MR, CR, SUBJ, MPQA, SST, TREC), one task on paraphrase detection (MRPC) and one on entailment classification (SICK-E). All these 8 tasks have accuracy as their performance measure. There are two tasks (SICK-R and STS14) for which the performance measure is Pearson and Pearson/Spearman correlation respectively.  The trained encoding models are used to generate initial representations for the sentences in the transfer tasks, which are then tuned further. For more details, please refer to the above paper. 

As can be seen from Table 1 and Fig. 2, among the models proposed in this paper, the \textsc{SufiSent-Tied} model with dimension 4096 has the best test accuracy on the SNLI dataset and also the best macro and micro average of the validation set accuracies in the 8 transfer tasks identified above. It also performs significantly better than the BiLSTM-Max (InferSent) of \cite{infersent}, which only uses the max of the prefix encodings in both directions. The performance steadily improves with increasing encoding dimension, as shown in Fig. 2. The test set performance of  \textsc{SufiSent-Tied}  on SNLI improves on InferSent too.  \textsc{SufiSent} and  \textsc{SufiSent-Tied} are close, with the latter edging forward in higher dimensions. 
\tabcolsep=0.08cm
\def\arraystretch{1.1}
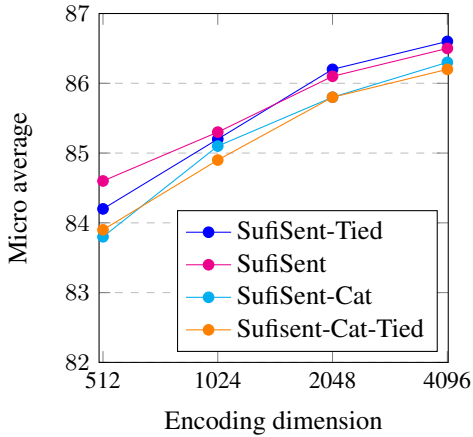
\begin{figure}[!t]
\begin{minipage}{\linewidth}
\begin{minipage}{0.44\linewidth}
\begin{tabular}{lccccc}
 & & \multicolumn{2}{ c}{\textbf{SNLI}} &  \multicolumn{2}{ c}{\textbf{Transfer}} \\
 
\textbf{Model} & \textbf{dim} & \textbf{dev} & \textbf{test} & \textbf{micro} & \textbf{macro}\\
\hline
BiLSTM-Mean & 4096 & 79.0 & 78.2 & 83.1 & 81.7\\
\hline
Inner-attention & 4096 & 82.3 & 82.5 & 82.1 & 81.0\\
\hline
HConvNet & 4096 & 83.7 & 83.4 & 82.0 & 80.9\\
\hline
BiLSTM-Max. & 4096  &  \textbf{85.0} & 84.5 & 85.2 & 83.7\\
\hline
\hline
SufiSent-Tied & 4096 & 84.7 & \textbf{84.6} &  \textbf{86.6} &  \textbf{85.1}\\
\hline
SufiSent & 4096 & 84.9 & 84.3 & 86.5 & 85.0\\
\hline
SufiSent-Cat-Tied & 4096 & 84.8 & 84.4 & 86.2 & 84.5\\
\hline
SufiSent-Cat & 4096 & 84.6 & 84.2 & 86.3 & 84.3\\
\hline
\end{tabular}

\medskip
Table 1. Performance of \textsc{SufiSent*} models on the SNLI dataset and on the validation 
sets of transfer tasks with accuracy as performance. Numbers in first four rows are taken from \cite{infersent}.\\
\end{minipage}
\hfill
\begin{minipage}{0.45\linewidth}
\begin{tikzpicture}
\begin{semilogxaxis}[
	width=0.99\linewidth,
       height=0.99\linewidth,
    xlabel={Encoding dimension },
    y label style={at={(axis description cs:0.1,.5)},anchor=south},
    ylabel={Micro average },
    xmin=500, xmax=4120,
    ymin=82, ymax=87.0,
    xtick={512, 1024, 2048, 4096},
    xticklabels={512, 1024, 2048, 4096},
    tick label style={font=\small},
    ytick={80, 81, 82, 83, 84, 85, 86, 87},
    legend pos=south east,
    ymajorgrids=true,
    grid style=dashed,
    legend cell align={left},
    legend entries={SufiSent-Tied, SufiSent, SufiSent-Cat, Sufisent-Cat-Tied}
]
 
\addplot[
    color=blue,
    mark=*,
    ]
    coordinates {
    (512, 84.2)(1024, 85.2)(2048, 86.2)(4098, 86.6)
    };
    \addplot[
    color=magenta,
    mark=*,
    ]
    coordinates {
    (512, 84.6)(1024, 85.3)(2048, 86.1)(4098, 86.5)
    };
    \addplot[
    color=cyan,
    mark=*,
    ]
    coordinates {
    (512, 83.8)(1024, 85.1)(2048, 85.8)(4098, 86.3)
    };
    
    \addplot[
    color=orange,
    mark=*,
    ]
    coordinates {
    (512, 83.9)(1024, 84.9)(2048, 85.8)(4098, 86.2)
    };

 \end{semilogxaxis}
\end{tikzpicture}
\captionof{figure}{Scaling of micro average of accuracies on 8 tasks with encoding dimension. }
\end{minipage}
\end{minipage}
\end{figure}
\begin{figure}[!t]
\begin{minipage} {\linewidth}
\tabcolsep=0.08cm
\def\arraystretch{1.1}
\begin{tabular}{lcccccccccc}
\hline
\textbf{Model}&\textbf{MR}&\textbf{CR}& \textbf{SUBJ} &  \textbf{MPQA} & \textbf{SST} & \textbf{TREC} & \textbf{MRPC} & \textbf{SICK-R} & \textbf{SICK-E} &\textbf{STS14}\\ 
\hline
\multicolumn{11}{ l}{\textit{Unsupervised Training - Ordered Sentences}} \\
\hline
FastSent               & 70.8 & 78.4 & 88.7 & 80.6 & - & 76.8 &  72.2/80.3 & - &  - &  {.63/.64} \\
\hline
SkipThought        & 76.5 & 80.1 & 93.6 & 87.1 & 82.0 & 92.2 & {73.0/82.0} & 0.858 & {82.3} & .29/.35 \\
\hline
SkipThought-LN  & {79.4} & {83.1} & {{93.7}} & {89.3} & {82.9} & {88.4} & -  &  {0.858} & 79.5 &.44/.45 \\
\hline
\hline
\multicolumn{11}{ l}{\textit{Supervised Training on SNLI}} \\
\hline
InferSent & 79.9 & 84.6 & 92.1 & 89.8 & 83.3 & {\textbf{88.7}} & 75.1/82.3 & 0.885 &  {\textbf{86.3}} & .68/.65 \\
\hline
SufiSent       & 80.3 & 84.7 & \textbf{92.8} & 90.1 & {\textbf{83.4}} & 88.0 & {\textbf{75.4/82.9}} & 0.886 & 85.7 &  {\textbf{.69/.66}} \\
\hline
SufiSent-Tied &  {\textbf{80.6}} &  {\textbf{85.4}} & 92.2 &  {\textbf{90.3}} & 83.1 & 88.4 & 74.3/82.3 &  {\textbf{0.887}} &  {\textbf{86.3}} & .68/.66 \\
\hline
SufiSent-Cat & 80.3 & 84.4 & 92.2 & 90.2 & 81.4 & 85.2 & 74.6/82.5 & 0.883 & 86.0 & .66/.63 \\
\hline
SufiSent-Cat-Tied & 79.8 & 84.8 & 92.3 & 90.2 & 82.3 & 86.6 & 74.5/82.5 & 0.880 & 85.8 & .64/.61\\
\hline \hline
\multicolumn{11}{ l}{\textit{Supervised Training on AllNLI}} \\
\hline
InferSent &  81.1 & 86.3 & 92.4 & 90.2 & 84.6 & 88.2 & 76.2/83.1 & 0.884 & 86.3  &.70/.67\\
\hline
\end{tabular}
\medskip

Table 2. Test set performance over the transfer tasks in SentEval. For MRPC, we report accuracy and F1 score. The dimension for the \textsc{SufiSent}* and Infersent models is 4096. All numbers except for our models are taken from \cite{hill2016} and  \cite{infersent}.  \\
\end{minipage}
\vspace{-0.7cm}
\end{figure}

Table 2 compares the test set performance of the \textsc{SufiSent} models with InferSent on each of the transfer tasks for the encoding dimension of 4096. For the same training set (SNLI),  both \textsc{SufiSent-Tied} and \textsc{SufiSent} improves or matches InferSent on 7 of the 10 tasks. The improvement is particularly significant for MR, CR, SUBJ and MPQA. The performance of models trained on unlabeled data and the InferSent model on the larger AllNLI dataset is also shown for comparison. 

To conclude, we propose \textsc{SufiSent} - a new universal sentence encoding that is computed by max-pooling over the encodings of the suffixes and prefixes of sentences, in both the forward and backward directions. Preliminary results obtained by training on the SNLI dataset shows promise, improving over existing approaches on many transfer tasks in the SentEval benchmark. In future work, we plan to train \textsc{SufiSent} on the larger AllNLI dataset, explore its use in other NLP tasks as a basic representation primitive and address computational efficiency issues.
\bibliography{bibtex}

\begin{thebibliography}{9}
\providecommand{\natexlab}[1]{#1}
\providecommand{\url}[1]{\texttt{#1}}
\expandafter\ifx\csname urlstyle\endcsname\relax
  \providecommand{\doi}[1]{doi: #1}\else
  \providecommand{\doi}{doi: \begingroup \urlstyle{rm}\Url}\fi

\bibitem[Bowman et~al.(2015)Bowman, Angeli, Potts, and Manning]{snli}
Samuel~R. Bowman, Gabor Angeli, Christopher Potts, and Christopher~D. Manning.
\newblock A large annotated corpus for learning natural language inference.
\newblock In \emph{EMNLP}, 2015.

\bibitem[Conneau et~al.(2017)Conneau, Kiela, Schwenk, Barrault, and
  Bordes]{infersent}
Alexis Conneau, Douwe Kiela, Holger Schwenk, Lo{\"i}c Barrault, and Antoine
  Bordes.
\newblock Supervised learning of universal sentence representations from
  natural language inference data.
\newblock In \emph{EMNLP}, 2017.

\bibitem[Deng et~al.(2009)Deng, Dong, Socher, Li, Li, and Li]{imagenet}
Jia Deng, Wei Dong, Richard Socher, Li-Jia Li, Kai Li, and Fei-Fei Li.
\newblock Imagenet: A large-scale hierarchical image database.
\newblock In \emph{CVPR}, 2009.

\bibitem[Hill et~al.(2016)Hill, Cho, and Korhonen]{hill2016}
Felix Hill, Kyunghyun Cho, and Anna Korhonen.
\newblock Learning distributed representations of sentences from unlabelled
  data.
\newblock In \emph{HLT-NAACL}, 2016.

\bibitem[Hochreiter \& Schmidhuber(1997)Hochreiter and Schmidhuber]{lstm}
Sepp Hochreiter and J{\"u}rgen Schmidhuber.
\newblock Long short-term memory.
\newblock \emph{Neural computation}, 9\penalty0 (8):\penalty0 1735--1780, 1997.

\bibitem[Kiros et~al.(2015)Kiros, Zhu, Salakhutdinov, Zemel, Urtasun, Torralba,
  and Fidler]{skipthought}
Ryan Kiros, Yukun Zhu, Ruslan~R Salakhutdinov, Richard Zemel, Raquel Urtasun,
  Antonio Torralba, and Sanja Fidler.
\newblock Skip-thought vectors.
\newblock In \emph{NIPS}, 2015.

\bibitem[Mikolov et~al.(2013)Mikolov, Sutskever, Chen, Corrado, and
  Dean]{word2vec}
Tomas Mikolov, Ilya Sutskever, Kai Chen, Greg Corrado, and Jeffrey Dean.
\newblock Distributed representations of words and phrases and their
  compositionality.
\newblock In \emph{NIPS}, 2013.

\bibitem[Mou et~al.(2016)Mou, Meng, Yan, Li, Xu, Zhang, and Jin]{mou2016}
Lili Mou, Zhao Meng, Rui Yan, Ge~Li, Yan Xu, Lu~Zhang, and Zhi Jin.
\newblock How transferable are neural networks in nlp applications?
\newblock \emph{CoRR}, abs/1603.06111, 2016.

\bibitem[Pennington et~al.(2014)Pennington, Socher, and Manning]{glove}
Jeffrey Pennington, Richard Socher, and Christopher~D Manning.
\newblock Glove: Global vectors for word representation.
\newblock In \emph{EMNLP}, 2014.

\end{thebibliography}
\bibliographystyle{iclr2018_workshop}

\end{document}